\documentclass[sigconf]{acmart}
\renewcommand\footnotetextcopyrightpermission[1]{} 

\usepackage{graphicx}
\usepackage{amsmath,amsfonts}
\usepackage{algorithmic}
\usepackage{textcomp}
\usepackage{multirow}

\usepackage{setspace}
\usepackage{amssymb}
\newcommand{\etal}{\textit{et al}.}
\newcommand{\ie}{\textit{i}.\textit{e}.}
\newcommand{\eg}{\textit{e}.\textit{g}.}

\AtBeginDocument{%
  \providecommand\BibTeX{{%
    \normalfont B\kern-0.5em{\scshape i\kern-0.25em b}\kern-0.8em\TeX}}}

\begin{document}
\fancyhead{}

\title{Cross-modality Discrepant Interaction Network for RGB-D Salient Object Detection}



\author{Chen Zhang$^{1,2}$, Runmin Cong$^{1,2,4,*}$, Qinwei Lin$^{1}$, Lin Ma$^{3}$, 
        Feng Li$^{1,2}$, Yao Zhao$^{1,2}$, Sam Kwong$^{4}$}
\makeatletter
\def\authornotetext#1{
\if@ACM@anonymous\else
    \g@addto@macro\@authornotes{\stepcounter{footnote}\footnotetext{#1}}
\fi}
\makeatother
\authornotetext{Corresponding author.}

\affiliation{
 \institution{\textsuperscript{\rm 1}Institute of Information Science, Beijing Jiaotong University, Beijing, China}
 \institution{\textsuperscript{\rm 2}Beijing Key Laboratory of Advanced Information Science and Network Technology, Beijing, China}
 \institution{\textsuperscript{\rm 3}Meituan, Beijing, China  \qquad $^4$City University of Hong Kong, Hong Kong, China}
 }
 
\email{{chen.zhang,rmcong,qinweilin,l1feng,yzhao}@bjtu.edu.cn, forest.linma@gmail.com, cssamk@cityu.edu.hk}
\def\authors{Chen Zhang, Runmin Cong, Qinwei Lin, Lin Ma, 
        Feng Li, Yao Zhao, Sam Kwong}
\renewcommand{\shortauthors}{Chen Zhang, et al.}


\begin{abstract}
\vspace{0.3cm}
The popularity and promotion of depth maps have brought new vigor and vitality into salient object detection (SOD), and a mass of RGB-D SOD algorithms have been proposed, mainly concentrating on how to better integrate cross-modality features from RGB image and depth map. For the cross-modality interaction in feature encoder, existing methods either indiscriminately treat RGB and depth modalities, or only habitually utilize depth cues as auxiliary information of the RGB branch. Different from them, we reconsider the status of two modalities and propose a novel Cross-modality Discrepant Interaction Network (CDINet) for RGB-D SOD, which differentially models the dependence of two modalities according to the feature representations of different layers. To this end, two components are designed to implement the effective cross-modality interaction: 1) the RGB-induced Detail Enhancement (RDE) module leverages RGB modality to enhance the details of the depth features in low-level encoder stage. 2) the Depth-induced Semantic Enhancement (DSE) module transfers the object positioning and internal consistency of depth features to the RGB branch in high-level encoder stage. Furthermore, we also design a Dense Decoding Reconstruction (DDR) structure, which constructs a semantic block by combining multi-level encoder features to upgrade the skip connection in the feature decoding. Extensive experiments on five benchmark datasets demonstrate that our network outperforms $15$ state-of-the-art methods both quantitatively and qualitatively. Our code is publicly available at:  \textit{\url{https://rmcong.github.io/proj_CDINet.html}.}

\end{abstract}





\maketitle

\section{Introduction}
\label{sec:introduction}
\vspace{0.3cm}
\begin{figure}[!t]
\centering
\centerline{\includegraphics[width=0.5\textwidth]{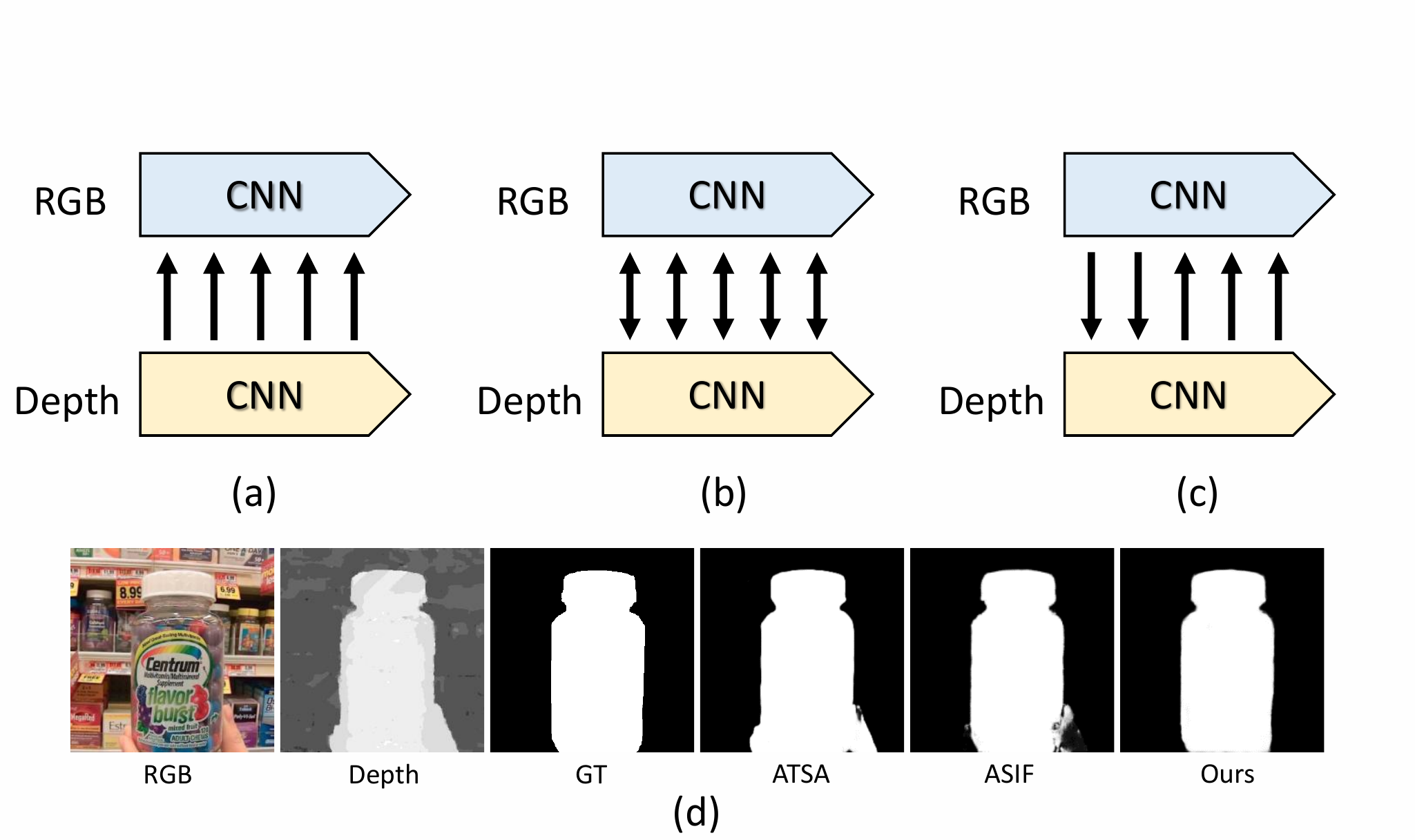}}

\caption{Mode (a) is a unidirectional interaction which regards the depth stream as auxiliary information; Mode (b) represents an undifferentiated bidirectional interaction, and it treats two modalities as equal; Mode (c) is the proposed mode which conducts discrepant cross-modality guidance.
In (d), ATSA~\cite{ATSA}, ASIF~\cite{ASIF} and Ours correspond to method of modes (a), (b) and (c) in a difficult scene, respectively.}
\label{fig1}
\end{figure}
Inspired by the human visual attention mechanism, salient object detection (SOD) aims to detect the most attractive objects or regions in a given scene, which has been successfully applied to abundant tasks~\cite{cong2018review,borji2015salient, cong2017co, cong2019video, DAFNet}.
In fact, in addition to the color appearance, texture detail, and physical size, people can also perceive the depth of field, thereby generating the stereo perception through the binocular vision system. In recent years, thanks to the rapid development of consumer depth cameras such as Microsoft Kinect, we are able to conveniently acquire the depth map to depict a scene. Compared with RGB image which provides rich color and texture information, depth map can exhibit the geometric structure, internal consistency, and illumination invariance. With the help of depth map, the SOD model can better cope with some challenging scenes, such as low-contrast and complex background. Therefore, for the past few years, the research on salient object detection of RGB-D images has received widespread attention. As we all know, RGB image and depth map belong to different modalities, 
thus we need some sophisticated designs to better use of the advantages of both to achieve RGB-D SOD. Therefore, the pivotal and hot issue in RGB-D SOD is how to better integrate cross-modality features. The limited expression ability of traditional models based on hand-crafted features~\cite{DCMC,LBE,Peng2014,NJUD,cong2019going,crm2019tmm,crm2019tc} makes their performance always unsatisfactory,  especially in complex scenes. With the popularity of deep learning in recent years, plenty of powerful cross-modality integration methods based on convolutional neural network (CNN) have been proposed~\cite{PGAR,JL-DCF,S2MA,HDFNet,MCI,cmMS,BiANet,SSF}.


For the information interaction between RGB and depth modalities in feature encoder, the existing mainstream methods can be roughly divided into two categories according to their interaction directions: (i) Unidirectional interaction mode shown in Figure \ref{fig1}(a), which uses the depth cues as auxiliary information to supplement the RGB branch~\cite{A2dele,PGAR,S2MA}. (ii) Undifferentiated bidirectional interaction shown in Figure \ref{fig1}(b), which treats RGB and depth cues equally to achieve cross-modality interaction~\cite{JL-DCF,HDFNet,MCI}. However, in this paper, we raise a new question: since these two modalities have their own strong points, can we design a discrepant interaction mode for RGB-D SOD based on their roles to make full use of the advantages of both? Observing the Figure~\ref{fig1}(d), we may be able to find the answer:
(1) Depth map has relatively distinct details (such as boundaries) for describing the salient objects, which is beneficial to straightforwardly learn the effective saliency-oriented features. However, when it comes to some special scenes, depth map can not distinguish different object instances at the same depth level only by virtue of its own characteristics, such as the bottle and fingers in Figure~\ref{fig1}(d).
At this point, the RGB branch can use its rich appearance detail information to enhance the depth feature learning.
(2) More affluent semantic information can be extracted from RGB image than the depth map, but the complex background interference or illumination variation influence may cause the salient objects to be flawed. By contrast, the depth features can provide better guidance for salient object positioning and internal consistency, thereby enhancing the semantic representation of the RGB modality.

Based on the above observations, we propose a cross-modality discrepant interaction network (CDINet) for RGB-D salient object detection, which clearly models the dependence of two modalities according to the manifestations of features in different layers.
Specifically, we first employ an RGB-induced detail enhancement (RDE) module in the first two layers of the encoding network, which can supplement more detailed information to enhance the low-level depth features.
Then, a depth-induced semantic enhancement (DSE) module is designed for high-level feature encoding, which utilizes the saliency positioning and internal consistency of high-level depth cues to enhance the RGB semantics. Thanks to this differentiated interaction mode, RGB and depth branches can complement each other, give full play to their respective advantages, and finally generate more accurate semantic representations.

In addition, for the encoder-decoder network, with the deepening of the convolution process, we can obtain global  semantic representation in the encoding stage, but some spatial details will be lost, thereby
only utilizing the supervision of ground truth in the decoder stage cannot achieve a perfect reconstruction result. In order to highlight and restore the spatial domain information in the feature decoding, the existing SOD models introduce the encoder features through skip connection \cite{HDFNet,S2MA}. However, they only introduce the information of the corresponding encoder layer through direct addition or concatenation operation, which do not make full use of the encoder features of different layers. To tackle this problem, we propose a dense decoding reconstruction (DDR) structure, which generates a semantic block by densely connecting the higher-level encoding features to provide more comprehensive semantic guidance for the skip connection in the feature decoding. Furthermore, we can obtain a more accurate and complete saliency prediction.
As can be seen from the visualization results in Figure \ref{fig1}(d), the ATSA~\cite{ATSA} method (unidirectional interaction) cannot well suppress the interference caused by the depth map, while the ASIF~\cite{ASIF} method ( undifferentiated bidirectional interaction) fails to completely detect the salient object. By contrast, our proposed CDINet can accurately detect the salient object with complete structure and clear background.

The main contributions of this paper are summarized as follows:
\begin{itemize}
  \item We propose an end-to-end Cross-modality Discrepant Interaction Network (CDINet), which differentially models the dependence of two modalities according to the feature representations of different layers. Our network achieves competitive performance against 15 state-of-the-art methods on 5 RGB-D SOD datasets. Moreover, the inference speed for an image reaches 42 FPS.
  \item We design an RGB-induced Detail Enhancement (RDE) module to transfer detail supplement information from RGB modality to depth modality in low-level encoder stage, and a Depth-induced Semantic Enhancement (DSE) module to assist RGB branch in capturing clearer and fine-grained semantic attributes by utilizing the advantage of positioning accuracy and internal consistency of high-level depth features.
  \item We design a Dense Decoding Reconstruction (DDR) structure, which generates a semantic block by leveraging multiple high-level encoder features to upgrade the skip connection in the feature decoding.
\end{itemize}

\section{Related Work}
\vspace{0.3cm}
\textbf{RGB-D Salient Object Detection.}
In recent years, a large number of RGB-D SOD models based CNN have shown extraordinary performance, and these works focus on how to design better integration strategies. However, most researchers are accustomed to treat the two modalities as equal, and we call them undifferentiated bidirectional interaction mode. For example, Fu \etal~\cite{JL-DCF} introduced a siamese network for joint learning and designed a densely-cooperative fusion strategy to discover complementary features. Pang \etal~\cite{HDFNet} integrated the cross-modal features through densely connected structure, then established a hierarchical dynamic filtering network by using fusion features. Huang \etal~\cite{MCI} proposed a cross-modal refinement module to integrate cross-modal features, then a multi-level fusion module was designed to fuse the features of each level followed bottom-up pathway.
In addition, some methods take depth map as auxiliary information for RGB branch, forming the unidirectional interaction mode. Piao \etal~\cite{A2dele} proposed a depth distiller to transfer the depth knowledge from depth stream to RGB stream.
Liu \etal~\cite{S2MA} designed a residual fusion module to integrate the depth decoding features into RGB branch in decoding stage. Chen \etal ~\cite{PGAR} considered depth map contains much less information than RGB image, then proposed a lightweight network to extract depth stream features.
Different from the above methods, we reconsider the status of RGB modality and depth modality in RGB-D SOD task, and propose a discrepant interaction structure to achieve more oriented and differentiated cross-modality interaction.


\textbf{Attention Mechanism.} As a form of efficient resource allocation, attention mechanism has been widely applied to plenty of computer vision tasks. Spatial attention mechanism~\cite{sa} makes the network pay attention to the area of interest. Channel attention mechanism~\cite{ca} learns the importance of each channel. Self-attention mechanism~\cite{non-local} captures long-distance dependency relationship. There are also several works that develop mixed attention mechanisms, such as CBAM~\cite{cbam} and dual-attention~\cite{dual}.
In this paper, we employ spatial-wise and channel-wise attention in RDE and DSE modules. Moreover, we focus more on the cross-modality application of attention, that is, the attention map generated by one modality is utilized to enhance another modality features, so as to achieve more effective cross-modality guidance in the form of attention.

\textbf{Skip Connection.}
Long-range skip connection is a measure to recover image details in pixel-level prediction tasks, and it has been equipped with almost all RGB-D SOD models. For models where cross-modal interaction occurs in encoder, skip connection is presented as a direct feature-wise addition or concatenation, \ie, \cite{S2MA} and \cite{HDFNet}. For other networks which fuse cross-modality features in decoder, proprietary modules are often designed to incorporate skip features (also known as side outputs). For example, Li \etal \cite{cmMS} proposed a cross-modality modulation and selection block to fuse side outputs in a coarse-to-fine way. Piao \etal \cite{DMRA} designed a depth refinement block to integrate complementary multi-level RGB and depth features. In this work, we boost performance by a simple but effective decoding structure that densely connects the higher-level encoder features to conduct skip connection.
\begin{figure*}[!t]
	\centering
	\includegraphics[width=1\linewidth]{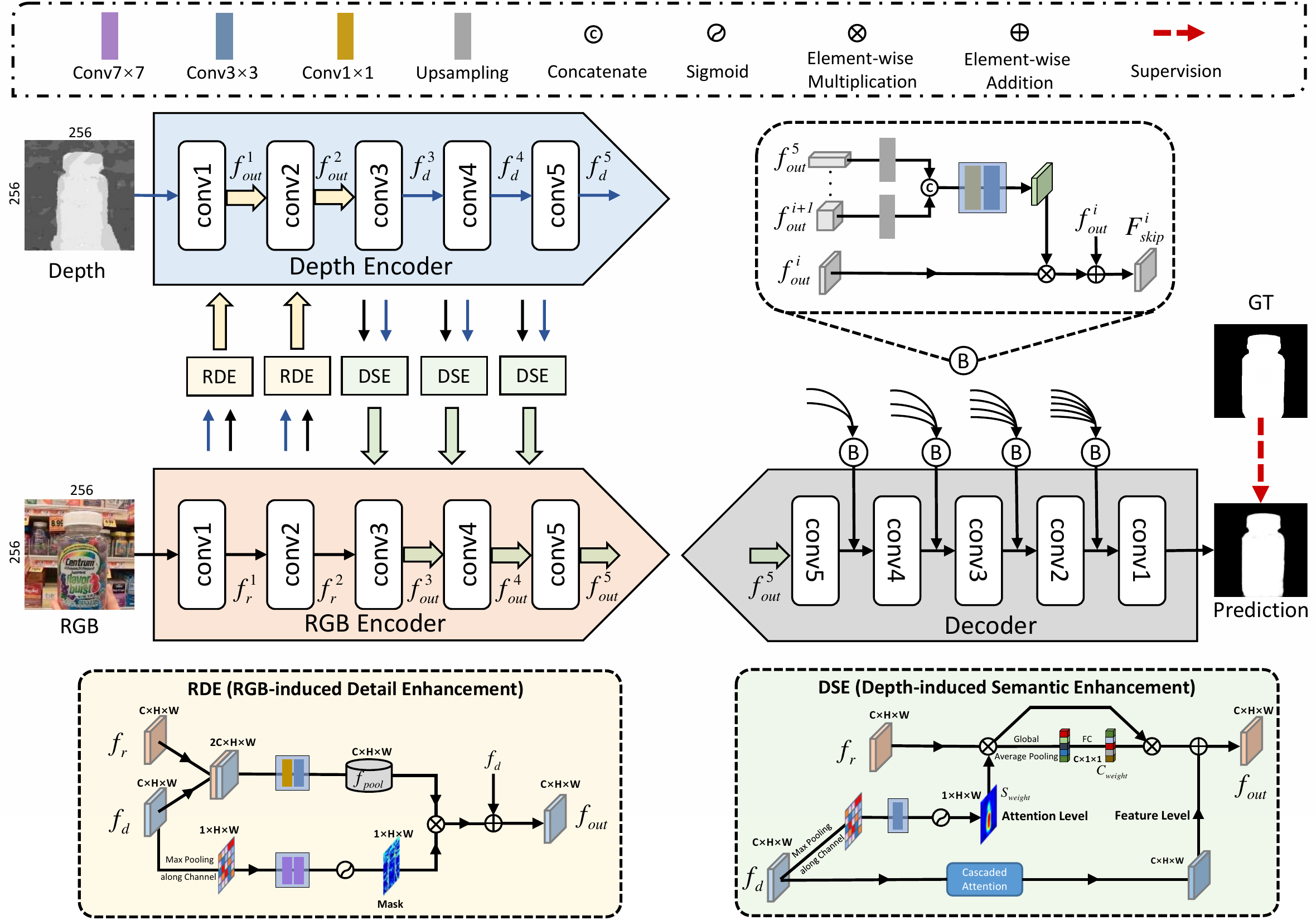}
	\caption{The overall pipeline of the proposed CDINet. Our CDINet follows an encoder-decoder architecture, which realizes the discrepant interaction and guidance of cross-modality information in the encoding stage. The framework mainly consists of three parts: 1) RGB-induced detail enhancement module. It achieves depth feature enhancement by transmitting the detailed supplementary information of the RGB modality to the depth modality. 2) Depth-induced semantic enhancement module. Depth features provide better positioning and internal consistency to enrich the semantic information of RGB features. 3) Dense decoding reconstruction structure. It densely encodes the encoder features of different layers to generate more valuable skip connection information, which is shown in the top right box marked B of this figure. The backbone of our network in this figure is VGG16~\cite{vgg}, and the overall network can be trained efficiently as an end-to-end system.}
	\label{framework}
\end{figure*}
\section{Proposed Method}
\vspace{0.3cm}
%

%
\subsection{Overview}
\vspace{0.2cm}
\textbf{Motivation.}
Previous studies have confirmed the positive effect of RGB and depth information interaction in SOD tasks \cite{S2MA,cmMS,ASIF}. In this paper, we seriously reconsider the status of RGB modality and depth modality in RGB-D SOD task. Different from the previous unidirectional interaction shown in Figure \ref{fig1}(a) and undifferentiated bidirectional interaction shown in Figure \ref{fig1}(b), we believe that the interaction of the two modalities information should be carried out in a separate and discrepant manner. The low-level RGB features can help the depth features to distinguish different object instances at the same depth level, while the high-level depth features can further enrich the RGB semantics and suppress background interference. Therefore, a perfect RGB-D SOD model should give full play to the advantages of each modality, and simultaneously utilize another modality to make up for itself to avoid causing interference. To this end, we propose a cross-modality discrepant interaction network to explicitly model the dependence of two modalities in the encoder stage according to the feature representations of different layers, which selectively utilizes RGB features to supplement the details for depth branch, and transfers the depth features to RGB modality to enrich the semantic representations.


\textbf{Architecture.}
Figure~\ref{framework} illustrates the overall architecture of the proposed CDINet for RGB-D SOD task. It is composed of three parts, \ie, the RGB-induced detail enhancement (RDE) module, the depth-induced semantic enhancement (DSE) module, and the dense decoding reconstruction (DDR) structure. On the whole, the network follows an encoder-decoder architecture, including two encoders regarding RGB and depth modalities and one decoder. The two encoders both adopt VGG16~\cite{vgg} network, discarding the last pooling and fully-connected layers, as the backbone to extract the corresponding multi-level feature representations and achieve cross-modality information interaction. The extracted RGB and depth features from the backbone are denoted as ${f}_{r}^i$ and ${f}_{d}^i$ respectively, where $r$ and $d$ represent the RGB and depth branches, and $i\in \{1,2,...,5\}$ indexes the feature level.
Specifically, in the low-level feature encoding stage (\ie, the first two layers of backbone), we design an RDE module to transfer detail supplement information from RGB modality to depth modality, thereby enhancing the distinguishability representation of depth features. For the high-level encoding features, the DSE module utilizes the advantage of positioning accuracy and internal consistency of depth features to assist RGB branch in capturing clearer and fine-grained semantic attributes, thereby promoting the object structure and background suppression.
Besides, for the convolution-upsample decoding infrastructure, we upgrade the traditional skip connection way by constructing a DDR structure, that is, utilizing higher-level skip connection features as guidance information to achieve more effective encoder information transmission. The prediction result generated by the last convolutional layer of decoder will be used as the final saliency output.
\subsection{RGB-induced Detail Enhancement}
\vspace{0.2cm}
\label{sec:RDE}
Compared with the RGB image, depth map puts aside complex texture information and can intuitively describe the shape and position of the salient objects. In this way, for the low-level encoder features that contain more detailed information (such as boundaries and shapes), depth features can provide more straightforward and instructive representations than RGB features, which are beneficial to the initial feature learning. However, depth information is not a panacea. For example, different object instances adjacent to each other have the same depth value, such as the bottle and fingers in Figure \ref{fig1}(d), which makes them manifest as indivisible objects in the depth map. But in the corresponding RGB image, these objects can be distinguished by the color difference in most cases. Hence, these ambiguous regions burden the network training, and previous models have confirmed the difficulty of predicting such samples. To address this dilemma, we design an RGB-induced detail enhancement module to reinforce and supplement depth modality through the RGB features in low-level layers. By introducing the detail guidance of RGB branch at an early stage, more information can be used in the feature feedforward process to handle these hard cases. The detailed architecture is shown in the bottom left of Figure \ref{framework}.


To be specific, we first adopt two cascaded convolutional layers to fuse the underlying visual features of two modalities. The first convolutional layer with the kernel size of $1\times1$ is used to reduce the number of feature channels, and the second convolutional layer with the kernel size of $3\times 3$ achieves more comprehensive feature fusion, thereby generating the fusion feature pool ${f}_{pool}$:
\begin{align}
{f}_{pool}^i=conv_3(conv_1([{f}_{r}^i, {f}_{d}^i])),
\end{align}
where $i\in \{1,2\}$ indexes the low-level encoder feature layer, $[\cdot,\cdot]$ denotes the channel-wise concatenation operation, and $conv_n(\cdot)$ is a convolutional layer with the kernel size of $n\times n$. The advantage of generating ${f}_{pool}$ instead of transferring the RGB features directly to the depth branch is that the common detail features of two modalities can be enhanced and irrelevant features can be weakened in this process.

Then, in order to cogently provide the useful information required by the depth features, we need to further filter the RGB features from the depth perspective. We use a series of operations on the depth features, including a maxpooling layer, two convolutional layers, and a sigmoid function, to generate a spatial attention mask as suggested by \cite{sa}. Note that for the two serial convolutional layers, we use a larger convolution kernel size (\ie, $7 \times 7 $) to perceive the important detail regions in a large receptive field. Finally, multiplying the mask and feature pool ${f}_{pool}$ to reduce the introduction of irrelevant RGB features, thereby obtaining the required supplement information from the perspective of depth modality. The entire process can be described as:
\begin{align}
{f}_{out}^i=\sigma(conv_7(conv_7(maxpool({f}_{d}^i))))\odot {f}_{pool}^i+{f}_{d}^i,
\end{align}
where $maxpool(\cdot)$ and $\sigma(\cdot)$ denote the maxpooling operation along channel dimension and sigmoid function respectively, and $\odot$ represents element-wise multiplication. The features ${f}_{out}^i$ ($i\in \{1,2\}$) will be used as the input of the next layer in depth branch. Note that, since the detail features in the depth branch are more intuitive and distinct, we choose them as skip connection features in the first two layers for decoding.

\subsection{Depth-induced Semantic Enhancement}
\label{sec:DSE}
\vspace{0.2cm}
In the high-level layers of encoder stage, the learned features of the network contain more semantic information, such as categories and relationships. For an RGB image, because it contains rich color appearance and texture content, its semantic information is also more comprehensive than depth modality. However, because of the relatively simple structure and data characteristics of the depth map, the learned high-level semantic features have better salient object positioning, especially in the suppression of the background regions, which is exactly what RGB high-level semantics require. Therefore, we design the depth-induced semantic enhancement module in the high-level encoder stage to enrich the RGB semantic features with the help of the depth modality. However, considering the simple fusion strategies (\eg, direct addition or concatenation) cannot effectively integrate cross-modality features. In this paper, we employ two types of interactive patterns to roundly carry out cross-modality features fusion, \ie, attention level and feature level. The detailed architecture is shown in the bottom right of Figure \ref{framework}.


First, we learn an attention vector $S_{weight}\in\mathbb{R}^{1\times h\times w}$ from the depth features to guide RGB modality to focus on the region of interest in a spatial attention~\cite{sa} manner, where the $h$ and $w$ represent the height and width of feature map, respectively. On the one hand, it helps to reinforce salient regions that are already recognized. On the other hand, it also allows the RGB branch to focus on information that is being ignored or incorrectly emphasized. This process can be formulated as:
\begin{align}
S_{weight}=\sigma(conv_3(maxpool({f}_{d}^i))),
\end{align}
\begin{align}
{f}_{rs}^i=S_{weight}\odot {f}_{r}^i,
\end{align}
where ${f}_{d}^i$ denotes the high-level encoder features of the depth branch, ${f}_{r}^i$ is the high-level RGB features from the backbone, ${f}_{rs}^i$ represents the RGB encoder features enhanced by the spatial attention of depth features ${f}_{d}^i$, and $i\in \{3,4,5\}$ indexes the high-level encoder feature layer. 
In addition, high-level features usually have abundant channels, so we use the channel attention~\cite{ca} to model the importance relationship of different channels and learn more discriminative features. Concretely, we learn the weight vector $C_{weight}\in\mathbb{R}^{c\times 1\times 1}$ through a global average pooling $(GAP)$ layer, two fully connected layers $(FC)$ and a sigmoid function, in which the $c$ denotes the number of channels in feature map. The final attention-level guidance is formulated as:
\begin{align}
C_{weight}=\sigma(FC(GAP({f}_{rs}^i))),
\end{align}
\begin{align}
D_{att}^i=C_{weight}\odot {f}_{rs}^i,
\end{align}
where $D_{att}^i$ denotes the attention-level RGB enhanced features.

As for guidance at the feature level, we use the pixel-wise addition operation to directly fuse the features of two modalities, which can strengthen the internal response of salient objects and obtain better internal consistency. It should be noted that we use cascaded channel attention \cite{ca} and spatial attention \cite{sa} mechanisms to enhance the depth features and produce the feature-level enhanced features $D_{add}^i$. Therefore, the features that eventually flow into the next layer of the RGB branch can be expressed as:
\begin{align}
& {f}_{out}^i=D_{att}^i+D_{add}^i.
\end{align}
Again, the enhanced features ${f}_{out}^i$ ($ i\in\{3,4,5\} $) of RGB branch will be introduced into the decoder stage to achieve saliency decoding reconstruction.
\subsection{Dense Decoding Reconstruction}
\label{sec:DFR}
\vspace{0.2cm}
In the feature encoding stage, we learn the multi-level discriminative features  through the discrepant guidance and interaction. The decoder is dedicated to learning the saliency-related features and predicting the full-resolution saliency map. During the feature decoding, skip connection that introduces encoding features into the decoder has been widely used in the existing SOD models~\cite{HDFNet,S2MA,DPANet,crm2019tgrs}. However, these methods only establish the relationship between the corresponding encoding and decoding layers, but ignore the different positive effects of different encoding features. For example, the top-layer encoding features can provide semantic guidance for each decoding layer. Consequently, we design a dense decoding reconstruction structure to more fully and comprehensively introduce skip connection guidance. In Figure~\ref{framework}, we show the specific implementation plan in the top right box.


To be specific, the ${f}_{out}^i$ of each layer in the encoding stage constitutes a skip connection features list. For convenience of distinction, we remark them as the skip connection features $f_{skip}^i~(i\in \{1, 2, 3, 4, 5\})$. Then, before the combination of decoding features and skip connection features of each layer, we densely connect the higher-level encoder features to generate a semantic block $B$, which is used to constrain the introduction of the skip connection information of the current corresponding encoder layer. The semantic block $B$ is defined as follows:
\begin{align}
B^i=conv_3(conv_1([up(f_{skip}^{i+1}),...,up(f_{skip}^{5})])),
\end{align}
where $up(\cdot)$ denotes the up-sampling operation via bilinear interpolation, which reshapes $f_{skip}^{j}(i<j\leq5)$ to same resolution with $f_{skip}^{i}$.

Then, with the semantic block, we adopt element-wise multiplication to eliminate redundant information, and a residual connection to preserve the original information, thereby generating the final skip connection features $F_{skip}^{i}$:
\begin{align}
F_{skip}^{i}=B^i\odot f_{skip}^{i}+f_{skip}^{i},
\end{align}
where $f_{skip}^{i}$ denotes the current corresponding skip connection features. In this dense way, the higher-level encoder features work as a semantic filter to achieve more effective information selection of skip connection features, thereby effectively suppressing redundant information that may cause anomalies in the final saliency prediction. The obtained $F_{skip}^i$ is combined with the decoding features of previous layer to gradually restore image details through up-sampling and successive convolution operations. Finally, the decoding features of the last layer are used to generate the predicted saliency map via a sigmoid activation.

\begin{figure*}[!t]
	\centering
	\includegraphics[width=1\linewidth]{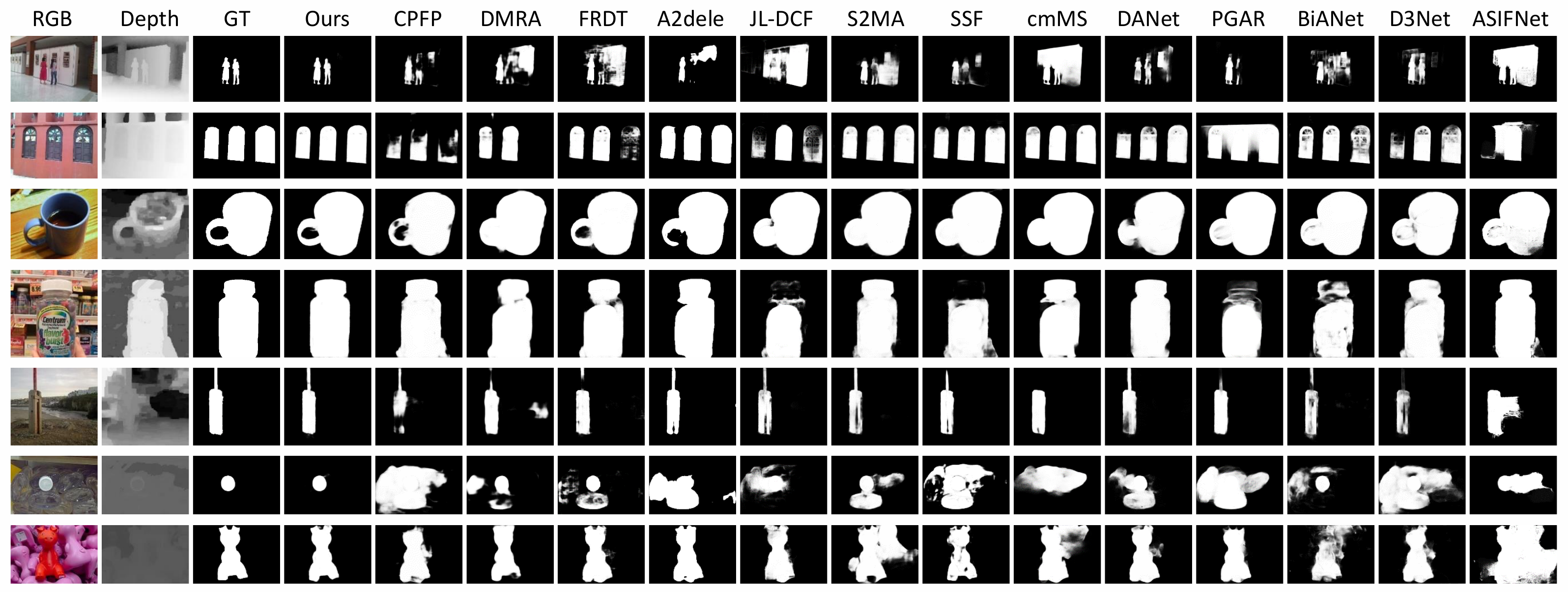}
	\vspace{-2em}
	\caption{Visual comparisons with other state-of-the-art RGB-D methods in some representative scenes.}
	\label{visualization}
\end{figure*} 
\section{Experiments}
\vspace{0.3cm}
\subsection{Datasets and Evaluation Metrics}
\vspace{0.2cm}
\noindent
\textbf{Benchmark Datasets.}
Five popular RGB-D SOD benchmark datasets are employed to evaluate the performance of the proposed model. The NLPR dataset~\cite{Peng2014} obtained by a Microsoft Kinect contains $1000$ pairs of RGB images and depth maps in indoor and outdoor locations. The NJUD dataset~\cite{NJUD} consists of $1985$ RGB images and corresponding depth maps, which are collected from 3D movies, the Internet, and stereo photographs. The DUT dataset \cite{DMRA} includes $1200$ indoor and outdoor complex scenes paired with corresponding depth maps. The STEREO dataset~\cite{Niu2012} collects $797$ stereoscopic images from Image gallery on the web, and obtains the corresponding depth maps through left-right view estimation. The LFSD dataset~\cite{LFSD} includes $100$ RGB-D images via a light field camera. Following~\cite{A2dele,cmMS}, we adopt $2985$ images as our training data, including $1485$ samples from NJUD, $700$ samples from NLPR, and $800$ samples from DUT, and all the remaining images are used as testing.

\begin{table*}[]
	\begin{center}
		\begin{spacing}{0.7}
			\small
			\caption{Quantitative comparison results in terms of S-measure ($S_\alpha$), max F-measure ($F_\beta$) and MAE score on five benchmark datasets. $\uparrow$ \& $\downarrow$ denote higher and lower is better, respectively. Bold number on each line represents the best performance.}
			\label{tab_compare}
			\vspace{0.15cm}
			\renewcommand\arraystretch{2}
			\setlength{\tabcolsep}{1.0mm}{
				\begin{tabular}{cc|ccccccccccccccc|c}
					\toprule[1.5pt]
					& \multirow{2}{*}{}
					& MMCI    & TAN  & CPFP   & DMRA  & FRDT  & SSF   & S2MA  & A2dele  & JL-DCF  & PGAR  & DANet   & cmMS   & BiANet  & D3Net & ASIFNet  & CDINet         \\ 
					&               & \cite{MMCI}  & \cite{TAN}  & \cite{CPFP}    & \cite{DMRA}
					& \cite{FRDT}   & \cite{SSF}    & \cite{S2MA}  & \cite{A2dele}   & \cite{JL-DCF}
					& \cite{PGAR}   & \cite{DANet}  & \cite{cmMS}  & \cite{BiANet}   & \cite{D3Net}
					& \cite{ASIF}   & Ours          \\  \cline{3-18}
					& \multirow{2}{*}{}
					& 2019    & 2019  & 2019   & 2019  & 2020  & 2020   & 2020  & 2020  & 2020  & 2020
					& 2020    & 2020  & 2020   & 2020  & 2021  & -      \\
					&         & PR    & TIP    & CVPR  & ICCV  & ACM MM & CVPR  & CVPR  & CVPR  & CVPR    & ECCV    & ECCV  & ECCV   & TIP   & TNNLS & TCyb   & -     \\  \hline
					\multirow{3}{*}{\rotatebox{90}{\textbf{NLPR}}}
					& $ F_{\beta}\uparrow$
					& .8149   & .8631 & .8675  & .8749 & .8976 & .8986  & .9017 & .8815 & .8915 & .9153
					& .9013   & .9031 & .8764  & .8969 & .8907 & \textbf{.9162}  \\
					& $ S_{\alpha}\uparrow$
					& .8557   & .8861 & .8884  & .8892 & .9129 & .9141  & .9155 & .8979 & .9097 &\textbf{.9297}   & .9152  & .9176 & .9000 & .9117  & .9079 & .9273  \\
					& $ MAE\downarrow$
					& .0591   & .0410 & .0359  & .0339 & .0290 & .0259  & .0298 & .0285 & .0295   & .0245 & .0283   & .0277 & .0325  & .0296 & .0295 & \textbf{.0240}  \\  \hline
					\multirow{3}{*}{\rotatebox{90}{\textbf{NJUD}}}
					& $ F_{\beta}\uparrow$
					& .8526   & .8741 & .7661  & .8883 & .8982 & .9000  & .8888 & .8733  & .9042  & .9068 & .8927   & .9034 & .9121  & .8996 & .8886 & \textbf{.9215}  \\
					& $ S_{\alpha}\uparrow$     & .8588   & .8785  & .7984 & .8804 & .8992 & .9002 & .8943 & .8704   & .9022   & .9089 & .8971   & .9051 & .9119   & .9002   & .8902     & \textbf{.9188}  \\
					& $ MAE\downarrow$           & .0789   & .0605 & .0794 & .0521 & .0467 & .0422 & .0532 & .0510   & .0413   & .0422 & .0463   & .0432 & .0399   & .0465   & .0472     & \textbf{.0354}  \\  \hline
					\multirow{3}{*}{\rotatebox{90}{\textbf{DUT}}}
					& $ F_{\beta}\uparrow$      & .7671   & .7903  & .7180 & .8975 & .9263 & .9242 & .8997 & .8923   & .8612   & .9171 & .8954   & .9090 & .8156   & .7855   & .8245     & \textbf{.9372}  \\
					& $ S_{\alpha}\uparrow$      & .7913   & .8083 & .7490 & .8879 & .9159 & .9157 & .9031 & .8864   & .8758   & .9136 & .8894   & .9070 & .8368   & .8152   & .8396     & \textbf{.9274}  \\
					& $ MAE\downarrow$           & .1126   & .0926 & .0955 & .0477 & .0362 & .0340 & .0440 & .0426   & .0556   & .0372 & .0465   & .0405 & .0745   & .0848   & .0724     & \textbf{.0302}  \\  \hline
					\multirow{3}{*}{\rotatebox{90}{\textbf{STEREO}}}
					& $ F_{\beta}\uparrow$       & .8425   & .8705 & .8601 & .8861 & .8987 & .8903 & .8158 & .8864   & .8740   & .9008 & .8199   & .8971 & .8844   & .8495   & .8800     & \textbf{.9033}  \\
					& $ S_{\alpha}\uparrow$     & .8559   & .8775  & .8714 & .8858 & .9004 & .8920 & .8424 & .8868   & .8855   & .9054 & .8410   & .8999 & .8882   & .8687   & .8820     &  \textbf{.9055} \\
					& $ MAE\downarrow$          & .0796   & .0591  & .0537 & .0474 & .0428 & .0449 & .0746 & .0431   & .0509   & .0422 & .0712   & .0429 & .0497   & .0578   & .0485     & \textbf{.0410}  \\  \hline
					\multirow{3}{*}{\rotatebox{90}{\textbf{LFSD}}}
					& $ F_{\beta}\uparrow$      & -       & -     & .8214 & .8523 & .8555 & .8626 & .8310 & .8280   & .8217   & .8390 & .8417   & .8623 & .7287   & .8062   & .8602     & \textbf{.8746}  \\
					& $ S_{\alpha}\uparrow$      & -       & -  & .8199   & .8393 & .8498 & .8495 & .8292 & .8258   & .8171   & .8444 & .8375   & .8491 & .7422   & .8167   & .8520     & \textbf{.8703}  \\
					& $ MAE\downarrow$           & -       & -  & .0953   & .0830 & .0809 & .0751 & .1018 & .0839   & .1031   & .0818 & .1031   & .0792 & .1340   & .1023   & .0809     &\textbf{.0631}   \\ 
					\bottomrule[1.5pt]
			\end{tabular}}
		\end{spacing}
	\end{center}
\end{table*}

\noindent
\textbf{Evaluation metrics.}
We adopt four commonly used metrics in SOD task to quantitatively evaluate the performance. P-R curve describes the relationship between precision and recall, and the closer to the upper right, the better the algorithm performance.

F-measure~\cite{Niu2012} indicates the weighted harmonic average of precision and recall by comparing the binary saliency map with ground truth:
\begin{align}
	&F_{\beta}=\frac{({\beta}^2+1) \cdot Precision \cdot Recall}{{\beta}^2 \cdot Precision+Recall},
\end{align}
where $ {\beta}^2 $ is set to $ 0.3 $ to emphasize the precision.

MAE score~\cite{cong2018review} calculates the difference pixel by pixel between the saliency map $S$ and ground truth $G$:
\begin{align}
	&MAE=\frac{1}{H\times W} \sum_{y=1}^{H}\sum_{x=1}^{W}\lvert S(x, y)-G(x, y) \rvert ,
\end{align}
where $H$ and $W$ are the height and width of the original image, respectively.

S-measure~\cite{fan2017structure} evaluates the object-aware ($ S_o $) and region-aware structural ($ S_r $) similarity between the predicted saliency map and ground truth, which is defined as:
\begin{align}
	&S=\alpha \ast S_o + (1-\alpha) \ast S_r,
\end{align}
where $ \alpha $ is set to $0.5$ for balancing the contributions of two terms. 
\subsection{Implementation Details}
\vspace{0.2cm}
We implement the proposed network using Pytorch framework and is accelerated by an NVIDIA GeForce RTX $2080$Ti GPU, and also implement our network by using the MindSpore Lite tool$\footnote{https://www.mindspore.cn/}$. All the training and testing images are resized to $256\times 256$, and the depth map is simply copied to three channels as input. During \textbf{training}, we initialize the parameters of backbone by the pre-trained model on ImageNet~\cite{imagenet}, and the other filters are initialized as the Pytorch default settings. Then, to avoid overfitting, we use random flipping and rotating to augment the training samples. Moreover, we apply the usual binary cross-entropy loss function to optimize the proposed network, and the Adam algorithm is used to optimize our network with the batch size of $4$ and the initial learning rate of 1e-4 which is divided by $5$ every $40$ epochs. The model can be trained in an end-to-end manner without any pre-processing (\eg, HHA~\cite{gupta2014learning} for depth map) or post-processing (\eg, CRF~\cite{krahenbuhl2011efficient}) techniques. It takes about $5$ hours to obtain the final model for $100$ epochs. When \textbf{testing}, the inference time for an image with size of $256\times 256$ is $0.023$ second (42 FPS) via the aforementioned GPU.

\subsection{Comparisons with SOTA Methods}
\vspace{0.1cm}
We compare our CDINet with $15$ state-of-the-art CNN-based RGB-D SOD methods, including MMCI~\cite{MMCI}, TAN~\cite{TAN},  CPFP~\cite{CPFP}, DMRA~\cite{DMRA}, FRDT~\cite{FRDT}, SSF~\cite{SSF}, S2MA~\cite{S2MA}, A2dele~\cite{A2dele}, JL-DCF~\cite{JL-DCF}, DANet~\cite{DANet}, PGAR~\cite{PGAR}, cmMS~\cite{cmMS}, BiANet~\cite{BiANet}, D3Net~\cite{D3Net}, and ASIFNet~\cite{ASIF}. For fair comparisons, we test these methods with the released codes under the default settings to obtain the saliency maps. As for the models without released codes, we directly use the saliency maps provided by the authors for comparison.
\\
\textbf{Quantitative evaluation.} Table~\ref{tab_compare} objectively indicates the quantitative comparison results in terms of three evaluation metrics on five datasets. It can be seen that our network outperforms all compared methods on these five datasets, except for the S-measure on the NLPR dataset. For example, compared with the \emph{second best} method on the DUT dataset, the minimum percentage gain reaches $1.2\%$ for max F-measure, $1.3\%$ for S-measure, and $11.2\%$ for MAE score. On the NJUD dataset, compared with the BiANet~\cite{BiANet} (the second best method), our proposed CDINet has a $1.0\%$ improvement for max F-measure, $0.8\%$ improvement for S-measure, and $11.3\%$ improvement for MAE score. On the LFSD dataset, compared with the \emph{second best} method, the S-measure is improved from 0.8520 to 0.8703, with the percentage gain of $2.1\%$. Limited by the page space, the P-R curves of different methods on five datasets are shown in the \emph{supplementary materials}. 

\noindent
\textbf{Qualitative comparison.} In order to more intuitively demonstrate the excellent performance of the proposed method, we provide some qualitative comparison results in Figure~\ref{visualization}. As we can see in this figure, our model achieves better visual effects in many challenging scenarios, such as small objects (\ie, the first image), multiple objects (\ie, the second image), and disturbing backgrounds (\ie, the fifth image).
Meanwhile, our method not only accurately detects salient objects, but also obtains better internal consistency. For example, in the second image, although other methods can detect multiple windows, they cannot guarantee good structural integrity and internal consistency of salient objects, while our CDINet does it, benefiting from the proposed DSE module. In addition, the dense decoding reconstruction structure also makes the decoding process more refined, and the object boundary is sharper (\eg, the last image). For the confusing depth map (\eg, the fourth image), most methods result in redundant areas or vague predictions, yet our method can effectively suppress these ambiguous regions.
%
\begin{table}[h]
	\begin{center}
		\begin{spacing}{0.7}
		\small
		\caption{Ablation analyses of different components on the NLPR and DUT datasets.}
		\label{tab_ab}
		\setlength{\abovecaptionskip}{0.cm}
		\setlength{\belowcaptionskip}{-0.5 cm}
		\renewcommand\arraystretch{1.5}
		\setlength{\tabcolsep}{2mm}{
			\begin{tabular}{c|c c c|c c c}
				\toprule[1.5pt]
				\multirow{2}{*}{models}
				& \multicolumn{3}{c|}{NLPR}
				& \multicolumn{3}{c}{DUT} \\ \cline{2-7}
				
				&$ F_{\beta}\uparrow$  & $S_{\alpha}\uparrow$  & $MAE\downarrow$
				& $ F_{\beta}\uparrow$  & $S_{\alpha}\uparrow$  & $MAE\downarrow$    \\ \hline
				\textbf{CDINet}  & \textbf{.9162}  & \textbf{.9273}
				& \textbf{.0240}  & \textbf{.9372}   & \textbf{.9274}  & \textbf{.0302}  \\
				 w/o RDE    & .9153   & .9261  & .0251  & .9327  & .9226  & .0338  \\
				 w/o DSE    & .9062   & .9219  & .0253  & .9222  & .9184  & .0369  \\
				 w/o DDR    & .9154   & .9258  & .0248  & .9296  & .9238  & .0334  \\
				\bottomrule[1.5pt]
		\end{tabular}}
		\end{spacing}
	\end{center}
\vspace{-0.2cm}
\end{table}

\subsection{Ablation Studies}
\label{sec:Ablation}
\vspace{0.2cm}
We conduct thorough ablation studies to analyze the effects of individual components in our CDINet on the NLPR and DUT datasets. The quantitative results are reported in Table~\ref{tab_ab}, in which the first line (\ie, CDINet) shows the performance of our full model. Moreover, we also explore the role of different interaction modes in Table~\ref{tab_an}.

\textbf{Effectiveness of RDE module.} First, we remove the RDE module to verify its role, denoted as 'w/o RDE', which means two modality information that does not interact in the first two layers. Compared with the full model listed in Table~\ref{tab_ab}, the performance is moderately enhanced after adding the RDE module, which achieves the percentage gain of $4.4\%$ and $10.7\%$ in terms of MAE score on the NLPR dataset and DUT dataset, respectively. This experiment shows the effectiveness of the RGB modality to guide the depth modality through the RDE module.

\begin{figure}[h]
\centering
\centerline{\includegraphics[width=0.45\textwidth]{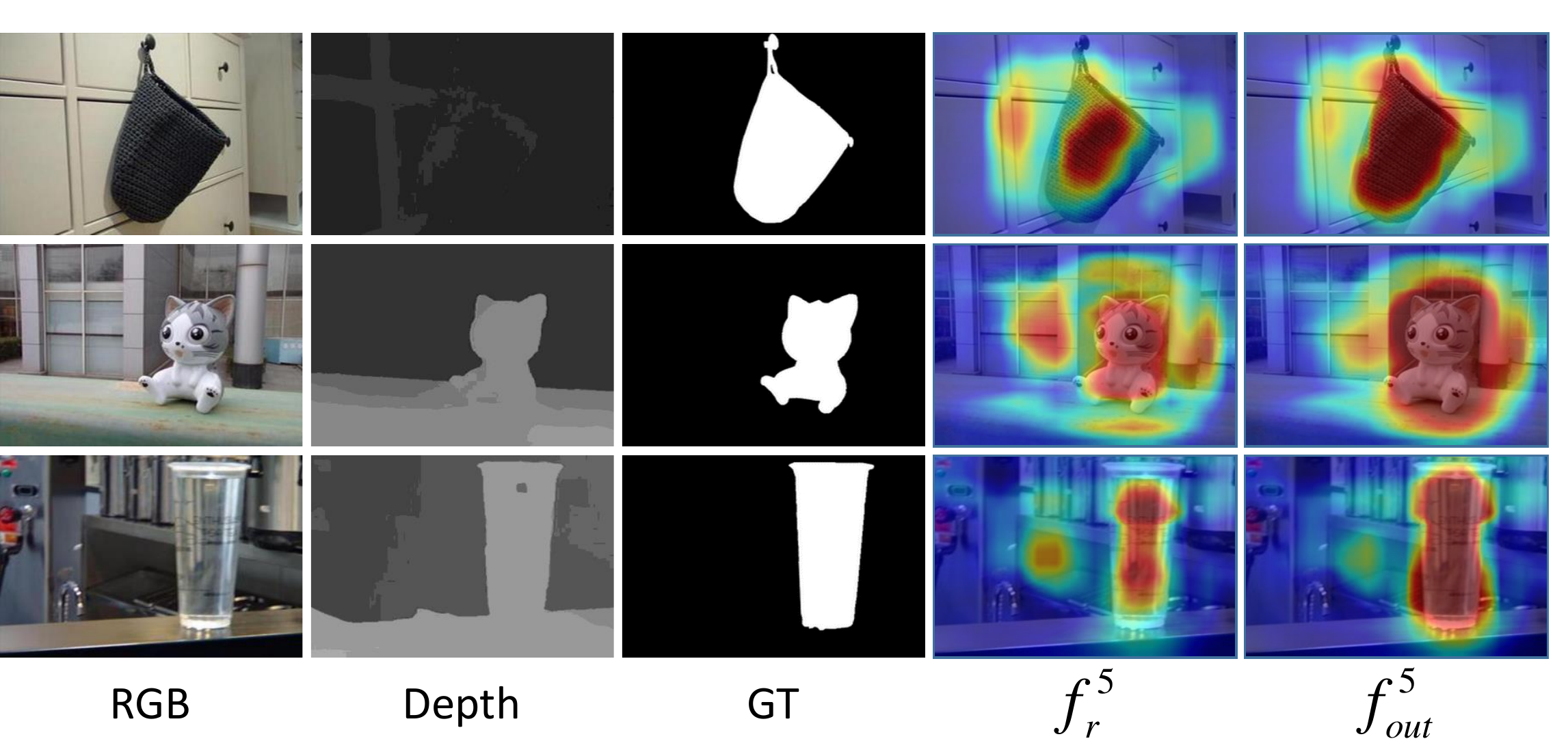}}
\caption{Feature visualization of the DSE module in the last layer of backbone.}
\label{DSE_vis}
\end{figure}

\textbf{Effectiveness of DSE module.} We also directly delete the DSE module in the $3^{rd}$, $4^{th}$ and $5^{th}$ encoders of two-stream backbone network, and then add the RGB features and depth features of the top layer for decoding. The results in the third row (\ie, 'w/o DSE') of Table \ref{tab_ab} demonstrate the positive effect of the DSE module. On the NLPR dataset and DUT dataset, without the DSE module, the max F-measure is decreased by $ 1.1\% $ and $ 1.6\% $, respectively. In addition, we show the visualization of the RGB features before and after DSE (\ie, ${f}_{r}^5$ and ${f}_{out}^5$) in Figure~\ref{DSE_vis}. 
With the introduction of depth guidance through the DSE module, the internal response of objects in the RGB features is obviously improved, while the abnormal noise in the background area is effectively suppressed.
We attribute the performance improvement to the attention-level and feature-level interactive patterns, which assists the RGB modality in capturing clearer and fine-grained semantic attributes.

\textbf{Effectiveness of DDR structure.} 
We replace the proposed dense decoding reconstruction structure by using a corresponding layer skip connection similar to U-net~\cite{UNet}, and the result is shown in the last line (\ie, w/o DDR) of Table~\ref{tab_ab}.
By comparing with the full network, it can be seen that our decoding strategy improves three metrics on two testing datasets, especially achieves the percentage gain of $ 0.8\% $ for max F-measure and $ 9.6\% $ for MAE score on the DUT dataset.
\begin{table}[!t]
	\begin{center}
	\begin{spacing}{0.7}
		\small
		\caption{The effectiveness analyses of discrepant interaction structure on the NLPR and DUT datasets.}
		\label{tab_an}
		\setlength{\abovecaptionskip}{0.cm}
		\setlength{\belowcaptionskip}{-0.5 cm}
		\renewcommand\arraystretch{1.5}
		\setlength{\tabcolsep}{2mm}{
			\begin{tabular}{c|c c c|c c c}
				\toprule[1.5pt]
				\multirow{2}{*}{Number}
				& \multicolumn{3}{c|}{NLPR}
				& \multicolumn{3}{c}{DUT} \\ \cline{2-7}
				
				& $ F_{\beta}\uparrow$  & $S_{\alpha}\uparrow$  & $MAE\downarrow$
				& $ F_{\beta}\uparrow$  & $S_{\alpha}\uparrow$  & $MAE\downarrow$    \\ \hline
				\textbf{No.1}     & \textbf{.9162}  & .9273
				& \textbf{.0240}  & \textbf{.9372}   & \textbf{.9274}  & \textbf{.0302}  \\
				No.2    & .9153   & .9261  & .0251  & .9295  & .9217  & .0345  \\
				No.3    & .9160   &\textbf{.9298}   & .0242  & .9328  & .9246  & .0327  \\
				\bottomrule[1.5pt]
		\end{tabular}}
		\end{spacing}
	\end{center}
\vspace{-0.2cm}
\end{table}

\textbf{Effectiveness of Discrepant Interaction.} In this work, we propose a novel interaction architecture, here, we conduct a couple of experiments to analyze the variants of interaction mode. 
First, we reset the guidance direction of the RSE module to the depth branch pointing to the RGB branch in the first two layers, forming the unidirectional interaction mode shown in Figure~\ref{fig1}(a).
As reported in the second line (denoted as 'No.2') of Table~\ref{tab_an}, compared with the full model (denoted as 'No.1'), the performance is obviously dropped on the DUT dataset. 
Furthermore, we also verify the bidirectional interaction mode shown in Figure~\ref{fig1}(b) by symmetrically inserting the RSE module and the DSE module into two branches. The results (denoted as 'No.3') show that almost all indicators of CDINet (\ie, No.1) are optimal, except for the S-measure on the DUT dataset. However, this performance improvement of the bidirectional interaction mode comes at the cost of increased computation and the number of parameters, compared with the final model, the parameter amount is increased by 10M and inference time changes from 42FPS to 32FPS. In general, our model achieves better results on the basis of considering both performance and computation cost.
\section{Conclusion}
\vspace{0.3cm}
In this paper, we explore a novel cross-modality interaction mode and propose a cross-modality discrepant interaction network, which explicitly models the dependence of two modalities in different convolutional layers. To this end, two components (\ie, RDE module and DSE module) are designed to achieve differentiated cross-modality guidance. Furthermore, we also put forward a DDR structure, which generates a semantic block by leveraging multiple high-level features to upgrade the skip connection. The comprehensive experiments demonstrate that our network achieves competitive performance against state-of-the-art methods on five benchmark datasets, and our inference speed reaches the real-time level (\ie, $42$ FPS).


\begin{acks}
\vspace{0.3cm}
This work was supported by the Beijing Nova Program under Grant Z201100006820016, in part by the National Key Research and Development of China under Grant 2018AAA0102100, in part by the National Natural Science Foundation of China under Grant 62002014, Grant U1936212, in part by Elite Scientist Sponsorship Program by the China Association for Science and Technology under Grant 2020QNRC001, in part by General Research Fund-Research Grants Council (GRF-RGC) under Grant 9042816 (CityU 11209819), Grant 9042958 (CityU 11203820), in part by Hong Kong Scholars Program under Grant XJ2020040, in part by CAAI-Huawei MindSpore Open Fund, and in part by China Postdoctoral Science Foundation under Grant 2020T130050, Grant 2019M660438.
\end{acks}


\bibliographystyle{ACM-Reference-Format}
\bibliography{sample-base}


\appendix
\title{Cross-modality Discrepant Interaction Network for RGB-D Salient Object Detection \textit{(Supplementary Material)}}
\maketitle
\section{More explanations}
\vspace{0.2cm}
\textbf{The design of RDE and DSE.}
Under the discrepant interaction mode, fully considering the role of different modalities, we design the RDE for low-level feature and DSE for high-level feature. (1) Considering that depth map has relatively clean details (such as boundary) for describing the salient objects, but cannot distinguish object instances at the same depth level, so the RDE is designed to transfer detail supplement information from RGB modality to depth modality. In implementation, we use a mask generated from depth feature to multiply fusion features in a gate manner, so that informative color information can be obtained from the perspective of depth feature. (2) Rich semantic information can be extracted from deeper RGB layers, but complex background may still blemish the positioning of salient objects. Hence, we design the DSE to transfer the object positioning and internal consistency of depth features to RGB. In implementation, a spatial attention map learned from depth feature is utilized to highlight the location of salient objects and suppress the background in RGB feature space. Simultaneously, the feature-level addition is used to enhance the internal consistency of salient objects with the guidance of depth feature. (3) These two modules are specially designed and are irreplaceable in terms of motivation and implementation. Here, to further prove the irreplaceability of RDE and DSE, we also implement lots of ablation studies: (a) If the RDE is replaced by DSE, it means that RGB feature needs to generate an attention map to weight the depth, and then perform feature-level addition. This is obviously unwise, because the self-highlighted features from the RGB branch cannot obtain the most valuable information that the depth branch really wants. The experiment also prove this point, the F-measure drops from 0.9162/0.9372 to 0.9142/0.9296 on the NLPR/DUT. (b) If the DSE is replaced by RDE, it means that RGB branch needs to explore information from the depth feature. However, what we want is to provide object positioning and internal consistency of depth feature for RGB branch. If it is replaced, it will hinder the role of the depth features. Under this setting, the F-measure are 0.9107 and 0.9261 on the NLPR and DUT. (c) Analogously, if the RDE and DSE are exchanged, the F-measure on the NLPR and DUT only reach 0.9066 and 0.9254. 

\textbf{The location of RDE and DSE. }
In the proposed discrepant guidance mode, we need to choose the appropriate information flow direction according to the feature representation of different layers. Therefore, following ~\cite{A2dele} and ~\cite{BiANet}, we regard the first two layers as low-level stage, and the last three layers as high-level stage. At the same time, related experiments have been carried out, for example,  when using RDE for the first three layers and DSE for the last two layers, the F-measure drops from 0.9162/0.9372 to 0.9138/0.9218 on the NLPR/DUT.

\section{Experiments}
\begin{figure*}[htbp]
	\centering
	\includegraphics[width=0.9\linewidth]{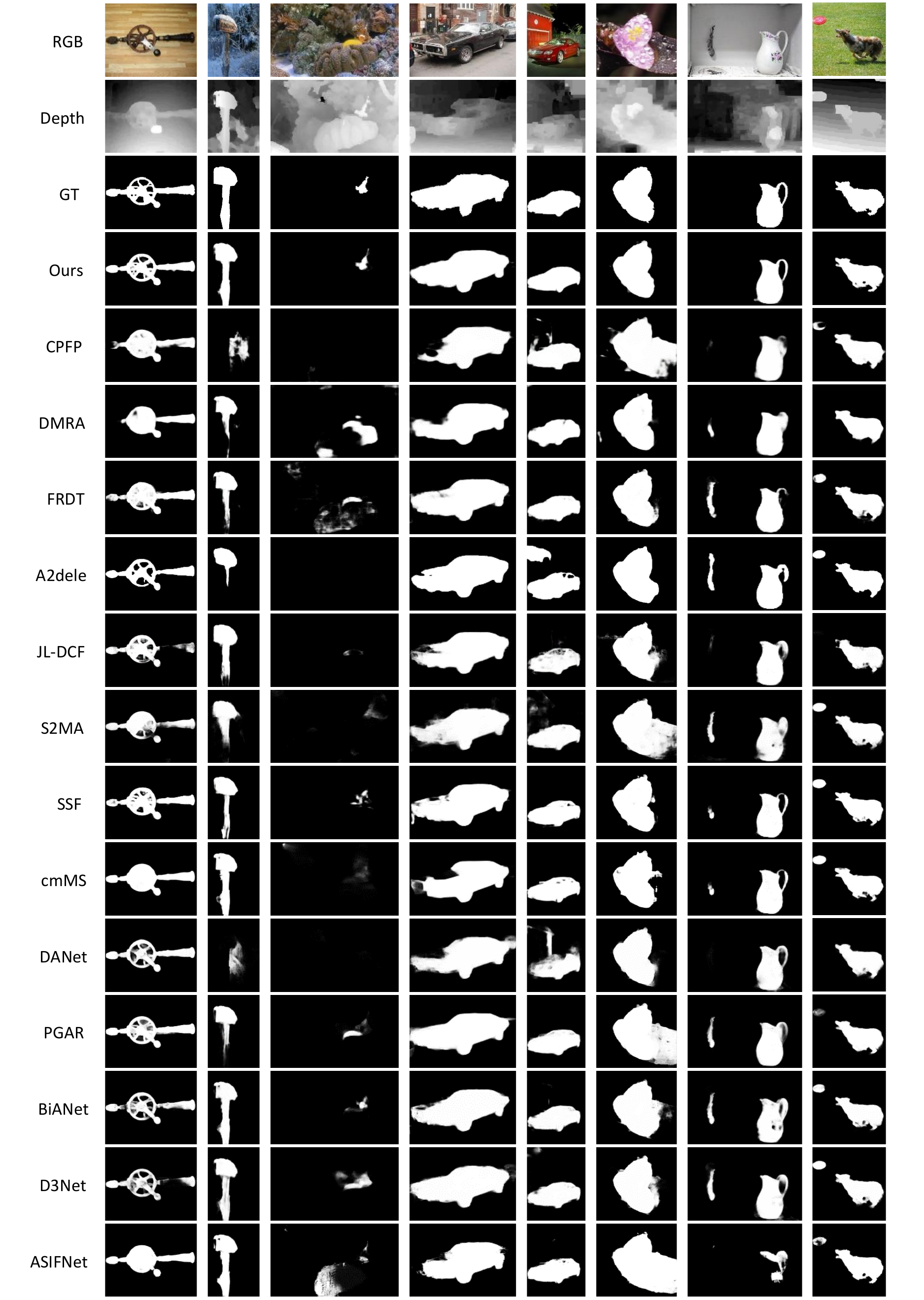}
	\caption{Visual comparisons with other state-of-the-art RGB-D methods in some representative scenes.}
	\label{vis_sup}
\end{figure*}

\begin{figure*}[h]
	\centering
	\includegraphics[width=1\linewidth]{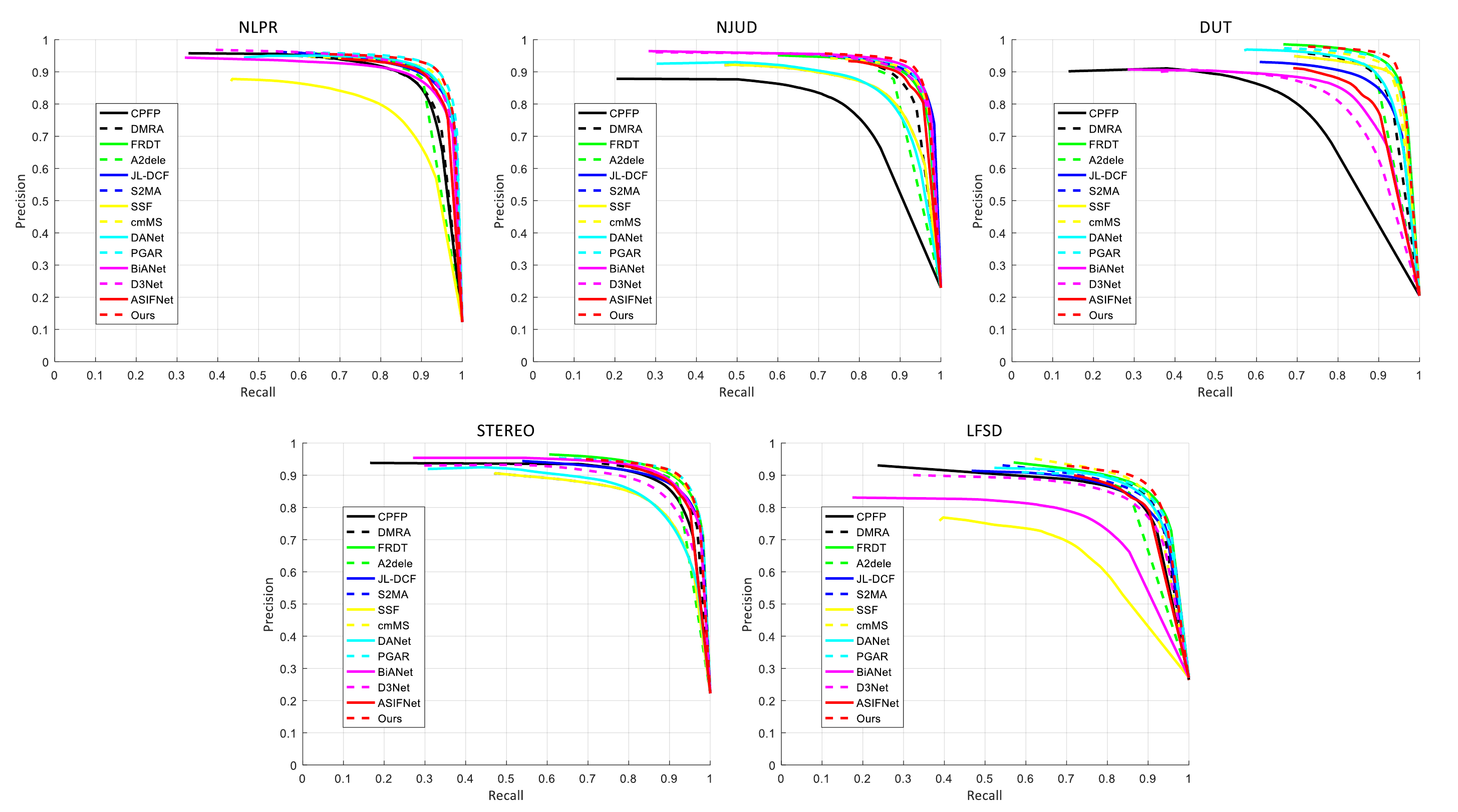}
	\caption{The P-R curves of different methods on five benchmark datasets.}
	\label{PR_curves}
\end{figure*}

\vspace{0.2cm}
We compare our CDINet with $15$ state-of-the-art CNN-based RGB-D SOD methods, including MMCI~\cite{MMCI}, TAN~\cite{TAN},  CPFP~\cite{CPFP}, DMRA~\cite{DMRA}, FRDT~\cite{FRDT}, SSF~\cite{SSF}, S2MA~\cite{S2MA}, A2dele~\cite{A2dele}, JL-DCF~\cite{JL-DCF}, DANet~\cite{DANet}, PGAR~\cite{PGAR}, cmMS~\cite{cmMS}, BiANet~\cite{BiANet}, D3Net~\cite{D3Net} and ASIFNet~\cite{ASIF} on five benchmark datasets (\ie, NLPR~\cite{Peng2014}, NJUD~\cite{NJUD}, DUT~\cite{DMRA}, STEREO~\cite{Niu2012}, and LFSD~\cite{LFSD}). In this \emph{supplementary materials}, we provide the more visual comparisons and P-R curves in Figure~\ref{vis_sup} and Figure~\ref{PR_curves}, respectively.

From Figure~\ref{vis_sup}, we can see that the proposed CDINet presents better visual effect in various challenging scenarios, such as fine object structure in the first column, disturbing background (the coral and other sea creatures) in the third column, easily mistaken objects in the last two columns (the pendant and frisbee).

P-R curve describes the relationship between precision and recall, and the closer to the upper right, the better the algorithm performance. As can be seen from Figure~\ref{PR_curves}, our model achieves both higher precision and recall scores against other methods on all testing datasets. 

\end{document}